# Modular Belief Updates and Confusion about Measures of Certainty in Artificial Intelligence Research[*]

Eric Horvitz and David Heckerman[**]

Medical Computer Science Group
Knowledge Systems Laboratory, TC135 SUMC, Stanford University

## Abstract

Over the last decade, there has been growing interest in the use of measures of *change* in belief for reasoning with uncertainty in artificial intelligence research. An important characteristic of several methodologies that reason with changes in belief or belief updates, is a property that we term *modularity*. We call updates that satisfy this property *modular updates*. Whereas probabilistic measures of belief update - which satisfy the modularity property were first discovered in the nineteenth century, knowledge and discussion of these quantities remains obscure in artificial intelligence research. We define modular updates and discuss their inappropriate use in two influential expert systems.

## Introduction

Most work in reasoning about uncertainty has centered on the manipulation of measures of *absolute* belief. That is, most methods for managing uncertainty concern themselves with questions of the form, Given current evidence, how certain is some hypothesis? Although more obscure, there has been investigation into the use of measures of *change* in belief. Here, questions are of the form, Given a piece of evidence, how has the certainty of some hypothesis changed? The distinction between reasoning with quantities representing changes in belief and reasoning with quantities representing absolute measures of belief is significant in the design and characterization of methods for handling uncertainty in reasoning systems.

In this paper, we focus on measures of change in belief that satisfy a special property that we call *modularity*. We will define modularity and describe quantities called *modular updates*. Modular updates are central to the intent of several methodologies for reasoning under uncertainty in expert systems. We will discuss two medical expert systems, MYCIN and INTERNIST-1 and show that both systems have implicitly assumed modularity in reasoning with uncertainty. Finally, we argue that MYCIN and INTERNIST-1 have used inappropriate measures of belief update.

## Definition of a modular update

In this section, we formalize the concept of a modular update. Before we examine the notion of modularity, let us define what we mean by an *update*. Informally, a belief update represents the change of belief in some hypothesis given a piece of evidence. This must be contrasted with the notion of absolute belief in a hypothesis given evidence. We shall assume in this paper, that belief is a continuous, ordered quantity. That is, it makes sense to talk about a *degree* of belief and to say that one belief is stronger than another. We insist that belief updates are also continuous and ordered. In addition, the combination of two updates which correspond to two different pieces of evidence should itself be an update. Before this basic update property is stated formally, it should be realized that updating a belief in a hypothesis H given some evidence E may depend on prior evidence for the hypothesis. Therefore, updates logically must have three arguments: a hypothesis H, new evidence E, and prior evidence e. With this, we formally capture the notion of an update by insisting that there is some function $\phi$ such that

$$U(H, E_1 E_2, e) = \phi[U(H, E_2, E_1 e), U(H, E_1, e)] \qquad (1)$$

where $U(H, E_1, e)$ denotes the update on H given $E_1$ and prior evidence e, $U(H, E_2, E_1 e)$ denotes the update on H given $E_2$ and prior evidence $E_1$ and e, and $U(H, E_1 E_2, e)$ denotes the combined update given prior evidence e.

Now we introduce the formal concept of modularity. We say that an update is modular, if the update is independent of prior evidence. Formally,

$$U(H, E, e) = U(H, E, \emptyset) \equiv U(H, E) \qquad (2)$$

We note that the notion of modularity is not new. For example, informal notions of modularity have been a central theme in the MYCIN expert system project[1].

Assuming modularity of belief updates may or may not be reasonable depending on the nature of dependencies within the domain of interest. However, the property is often assumed in expert system research to eliminate the need to consider possible complex interactions among evidence. The modularity property thereby eases the task of knowledge acquisition and explanation, facilitates the construction and maintenance of knowledge bases and allows the use of relatively simple functions for belief updating. The desirability of these features makes it tempting to use the modularity assumption even when it is inappropriate.

## History and derivation of a modular update

The first modular update was proposed in 1878 by Peirce[2]. This quantity, which can be called "weight of evidence," is equal to the logarithm of a ratio of conditional probabilities commonly called the *likelihood ratio*. Several other researchers independently discovered this simple but powerful measure of change in belief. They include Turing[3], Good[4] and Minsky and Selfridge[5]. To see that this quantity is a modular update, first consider Bayes' Theorem for evidence E and hypothesis H:

$$p(H|Ee) = \frac{p(E|He)p(H|e)}{p(E|e)}$$

[*]This work was supported in part by the Josiah Macy, Jr. Foundation and the Henry J. Kaiser Family Foundation.

[**]Order of authorship based on a coin flip.

The corresponding formula for the negation of the hypothesis, $\neg H$, is

$$p(\neg H|Ee) = \frac{p(E|\neg He)p(\neg H|e)}{p(E|e)} \quad (3)$$

Dividing the two, we get

$$\frac{p(H|Ee)}{p(\neg H|Ee)} = \frac{p(E|He)}{p(E|\neg He)} \frac{p(H|e)}{p(\neg H|e)} \quad (4)$$

The quantity

$$\frac{p(E|He)}{p(E|\neg He)}$$

is called the likelihood ratio, and is written $\lambda(H,E,e)$. To see that $\lambda$ is an update by definition (1), consider Bayes' theorem for two pieces of evidence:

$$\frac{p(H|E_1E_2)}{p(\neg H|E_1E_2)} = \frac{p(E_1E_2|H)}{p(E_1E_2|\neg H)} \frac{p(H)}{p(\neg H)} = \frac{p(E_2|HE_1)}{p(E_2|\neg HE_1)} \frac{p(E_1|H)}{p(E_1|\neg H)} \frac{p(H)}{p(\neg H)} \quad (5)$$

Therefore, by the definition of $\lambda$ we see that

$$\lambda(H,E_1E_2,e) = \lambda(H,E_2,E_1e) \, \lambda(H,E_1,e). \quad (6)$$

In particular, $\lambda$ is a probabilistic update with the update combination function, $\phi$, being simple multiplication.

If we assume that evidence for a hypothesis is conditionally independent on the hypothesis and its negation, then

$$\lambda(H,E,e) = \lambda(H,E,\emptyset). \quad (7)$$

In other words, the likelihood ratio is a modular update when conditional independence is assumed.

Note that with this assumption, the combination of the update $\log[\lambda]$ takes on the simple form

$$\log[\lambda(H,E_1\ldots E_n)] = \Sigma_i \log[\lambda(H,E_i)].$$

Peirce referred to these modular updates as "weights of evidence" because they are added just as weights on a scale. Peirce argued that $\log[\lambda]$ is a useful probabilistic quantity for acquiring and making inferences with uncertainty.

### Historical confusion about absolute belief and belief updates

The use of well-characterized updates in reasoning systems has been rare. Only a few systems have accurately manipulated well-understood measures of change in belief (e.g., the GLASGOW DYSPEPSIA system, using the measure in $\lambda[(H,E)]$[5]). One explanation for the infrequent and often inappropriate use of measures of change in belief is that discussion of the quantities is relatively obscure; treatises on probability theory often focus solely on the use of absolute measures of belief in inferential reasoning. Another reason may be related to past confusion about the relationship of belief updates and absolute belief in the literature.

Confusion surrounding the relationship of measures of absolute belief and belief updates surfaced in discussions in the mid-twentieth century as researchers began to rediscover the subjective interpretation of probabilities. Whereas Bayes proposed probability as a personal measure of belief over two hundred years ago, the early twentieth century was marked by the rise of the interpretation of probabilities as frequencies. Although the frequency interpretation of probability is still quite popular, the early 1930's saw a "rediscovery" of probability as a measure of personal belief[7]. Among others, Ramsey, Savage, Carnap, and DiFenetti led this rediscovery.

Carnap originally introduced the phrase, *degree of confirmation* in his classic work, *Logical Foundations of Probability*[8], primarily to distinguish the subjective interpretation of probabilities from the classical frequency interpretation. While Carnap's original intent was noble, he and others began to use degree of confirmation to represent two very different concepts. Whereas he used degree of confirmation to denote what is now commonly called subjective probability, he also used the phrase to refer to change in belief. Meanwhile, within the philosophy of science literature, Popper[9] introduced *corroboration*, his term for a clearly defined measure of change in belief based on a combination of probabilistic quantities. In trying to disentangle terminology about measures of change in belief, Popper pointed out problems with Carnap's inconsistent use of the word confirmation. Much confusion had already proliferated by the time Carnap apologized for his wavering use of terminology in the second of edition of his book[10].

We note that the quantities defined by Popper are very similar to Peirce's "weights of evidence." In fact, in 1960, Good showed that the log-likelihood ratio satisfies a slightly modified version of Popper's axioms of corroboration[11].

We believe that confusion in the mid-twentieth century about terms such as confirmation and corroboration and poor understanding of their relevance to measures of change in belief might indeed be contributing factors to more recent confusion surrounding the use of modular belief updates versus absolute belief in machine intelligence research.

### Modular updates in MYCIN

MYCIN, EMYCIN, and its descendants have reasoned with measures of changes in belief called certainty factors (CF's). MYCIN's knowledge is represented as *rules* in the form of IF E THEN H where H is a hypothesis and E is evidence having relevance to the hypothesis. A certainty factor is associated with each MYCIN rule. CF's were designed as measures of *change* in belief about a hypothesis given some evidence[12]. The quantities range between -1 and 1; positive numbers correspond to an *increase* in belief in a hypothesis while negative quantities correspond to a *decrease* in belief when certain evidence becomes available.

Often in MYCIN, several pieces of evidence are relevant to the same hypotheses. Thus, the creators of the CF model developed methods for *combining* certainty factors. In the original work, they presented a function for calculating an effective certainty factor for two pieces of evidence in terms of certainty factors for each piece of evidence separately. That is, they assumed the basic update property, (1). Furthermore, they implicitly assumed that the CF quantity satisfied the modularity property (2), by writing certainty factors as a function of only H and E.

Now let us consider the original *definition* of certainty factors:

$$CF(H,E) = \begin{cases} \dfrac{p(H|E) - p(H)}{1 - p(H)} & p(H|E) > p(H) \\[2ex] \dfrac{p(H|E) - p(H)}{p(H)} & p(H) > p(H|E) \end{cases} \quad (8)$$

This definition clearly demonstrates that certainty factors are intended to represent changes in belief. However, there are problems with the definition. First, we shall show that making the assumption of conditional independence of evidence given H and ~H is inconsistent with the modularity axiom. We find this inconsistency to be unreasonable as it is impossible to capture complex interactions among evidence for hypotheses when modularity is assumed. Second, we shall show that, in certain situations, the modularity axiom implies *marginal independence* which is unacceptable as it makes updating impossible.

Consider the natural extension of the certainty factor definition to include prior evidence:

$$CF(H,E,e) = \begin{cases} \dfrac{p(H|Ee) - p(H|e)}{1 - p(H|e)} & p(H|Ee) > p(H|e) \\ \\ \dfrac{p(H|Ee) - p(H|e)}{p(H|e)} & p(H|e) > p(H|Ee) \end{cases} \quad (9)$$

Under the assumption of conditional independence of evidence given H and ~H, it can be shown by counterexample that the above CF definition does not satisfy the modularity property, (2). In particular, consider updates where

$p(H) = .01$,

$\lambda(H,E_1) = 99$, and

$\lambda(H,E_2) = .99$

In this case it is easy to show that

$CF(H,E_1,\emptyset) \approx .5$ and

$CF(H,E_1,E_2) \approx 1$

Indeed, assuming conditional independence leads to a violation of the modularity axiom.

We now consider the definition of certainty factors in a limiting case. In fact, this situation was discussed in the original work in a section concerning the elicitation of certainty factors from experts[12]:

> We noted earlier that experts are often willing to state degrees of belief in terms of conditional probabilities [p(H|E)] .... It is perhaps revealing to note, therefore, that when the *a priori* belief is small (i.e., p(H) is close to zero), the CF of a hypothesis confirmed by evidence is approximately equal to its conditional probability on that evidence:
>
> $$CF(H,E) = \frac{p(H|E) - p(H)}{1 - p(H)} \approx p(H|E)$$
>
> .... This observation suggests that confirmation, to the extent that it is adequately represented by CF's, is close to conditional probability (in certain cases), although it still defies analysis as a probability measure.

The authors are saying that the posterior probability of H given E is approximately equal to the certainty factor for H and E (a modular update) when p(H) is small. Note that if this argument is applied to the definition of certainty factors extended to include prior evidence, we get

$CF(H,E,e) \approx p(H|Ee)$ when $p(H|e) \approx 0$.

However, (2) with $E \equiv \emptyset$ then implies that

$p(H|e) = p(H)$.

That is, evidence and hypotheses are marginally independent. Using the limiting value of the CF definition as an approximation for a modular belief update implies that evidence can have no impact on belief in hypotheses. Given such a result, it is surprising that the original definition of certainty factors was not questioned. Unfortunately, in the case of MYCIN, this problem led to the elicitation of absolute quantities in lieu of updates[13]. We note here that it is possible to find updates which are functions of the prior and posterior probabilities, in the spirit of (8), which are modular. These quantities are closely related to the likelihood ratio $\lambda$[14].

## Modular updates in INTERNIST-1

INTERNIST-1 was the result of a major research project to build an expert system for internal medicine diagnosis. The system uses an *ad hoc* methodology for reasoning under uncertainty. The method for managing uncertainty in INTERNIST-1 is associated with similar problems with modularity. In fact, INTERNIST-1, like MYCIN, uses quantities elicited as absolute measures of belief in a methodology that implicitly assumes they are modular updates.

The creators of INTERNIST-1 elicited measures of belief from experts called *evoking strengths*. Evoking strengths can take on integer values between 0 and 5. An informal definition is associated with each of these values. For example, an evoking strength of 1 means that the "diagnosis [of the disease in question] is a rare or unusual cause of the listed manifestation." An evoking strength of 3 means that the "diagnosis is the most common but not the overwhelming cause of listed manifestation." And an evoking strength of 5 means that the "listed manifestation is pathognomonic for [implicates] the disease."

Evoking strengths are elicited in quite the same way as one would elicit the posterior probability p(H|E). The measure is viewed as an answer to the question "Given a patient with this finding, how strongly should I consider this diagnosis to be its explanation?"[15] In fact, evoking strengths are described as being "somewhat analogous to a posterior probability."[16]

In INTERNIST-1, evoking strengths are used in determining the "scores" for disease hypotheses. A disease's score, in turn, roughly reflects the belief that it is present in a patient; a high score is associated with a likely disease hypothesis. The score for a disease is computed in part by summing evoking strengths. This strongly suggests that these quantities are intended to be updates as defined by the basic update property (1). Furthermore, evoking strengths are only a function of the two arguments H and E. Therefore, INTERNIST-1 researchers also implicitly assumed that evoking strengths satisfy the modularity property (2). However, as in MYCIN, the assumption that p(H|Ee) is a modular update leads to the unacceptable result that evidence and hypotheses are marginally independent.

## Summary and conclusions

We have formally defined the property of modularity with respect to measures of change in belief, which we call *modular updates*. As we have discussed, modular updates were introduced in the nineteenth century but have remained obscure in comparison with the more familiar use of absolute measures of belief in reasoning under uncertainty. We have argued that two plausible reasoning systems have used inappropriate measures of belief in systems that assume the modularity property. It is important that measures of belief or belief update are combined in a way consistent with their elicitation from experts lest serious flaws in reasoning may result. We find confusion about modular updates in two influential systems to be troubling. We believe that future systems would benefit from explicit consideration of the properties and requirements associated with the use of modular belief updates in plausible reasoning.

## Acknowledgements

We would like to thank Curt Langlotz, Mark Musen, Lawrence Fagan and Ted Shortliffe for useful discussions.